\documentclass[conference]{IEEEtran}
\IEEEoverridecommandlockouts
\usepackage{cite}
\usepackage{amsmath,amssymb,amsfonts}
\usepackage{algorithmic}
\usepackage{graphicx}
\usepackage{textcomp}
\usepackage{xcolor}
\usepackage{booktabs} 
\usepackage{multirow}
\usepackage{tikz}
\usepackage{array} 
\usepackage{xcolor} 
\usepackage{amsmath}
\usepackage{csquotes}
\def\BibTeX{{\rm B\kern-.05em{\sc i\kern-.025em b}\kern-.08em
    T\kern-.1667em\lower.7ex\hbox{E}\kern-.125emX}}

\usepackage{geometry}
\begin{document}

\title{Systematic Evaluation of Novel View Synthesis\\ for Video Place Recognition}

\author{
\IEEEauthorblockN{Muhammad Zawad Mahmud, Samiha Islam, Damian Lyons} 
\IEEEauthorblockA{
Dept. of Computer \& Information Science\\
Fordham University, NY USA\\
\{mmahmud9,sislam77,dlyons\}@fordham.edu 
}
}

\maketitle
\begin{tikzpicture}[remember picture, overlay]
    \node[anchor=north west, xshift=1.5cm, yshift=-0.5cm] at (current page.north west)
    {This work has been submitted to the IEEE for possible publication.};
\end{tikzpicture}
\begin{tikzpicture}[remember picture, overlay]
    \node[anchor=north west, xshift=1.5cm, yshift=-1cm] at (current page.north west)
    {Copyright may be transferred without notice, after which this version may no longer be accessible.};
\end{tikzpicture}

\begin{abstract}
The generation of synthetic novel views has the potential to positively impact robot navigation in several ways. 
In image-based navigation, a novel overhead view generated from a scene taken by a ground robot could be used to guide an aerial robot to that location. In Video Place Recognition (VPR), novel views of ground locations from the air can be added that enable a UAV to identify places seen by the ground robot, and similarly, overhead views can be used to generate novel ground views.

This paper presents a systematic evaluation of synthetic novel views in VPR using five public VPR image databases and seven typical image similarity methods. We show that for small synthetic additions, novel views improve VPR recognition statistics. We find that for larger additions, the magnitude of viewpoint change is less important than the number of views added and the type of imagery in the dataset.
\end{abstract}


\section{Introduction}

Recent progress in generative visual artificial intelligence (AI) \cite{seo2024genwarp,flynn2016deepstereo,mildenhall2021nerf} offers a potential advantage for visual navigation, that is, the generation of realistic expected visual imagery based on the current camera image. In the case of a navigation system comprising ground robots and aerial robots, cross-view registration requires a feature or semantic mapping process \cite{Miller_2022}. This could be simplified if generative visual AI techniques can use the aerial view to generate a corresponding ground view (or vice versa). For a team of robots using a group visual homing approach \cite{lyonsrahouti2023}, a ground robot image of a target can be used to generate a corresponding aerial view, allowing an aerial robot to visually home on the target. However, the core question with this approach is whether the novel view synthesized from the camera image corresponds sufficiently with the unseen visual imagery to be considered a useful view of the same physical location. We evaluate this question here using a Visual Place Recognition (VPR) framework\cite{Schubert2023VisualPR}.

VPR involves comparing a current robot camera image with previously recorded imagery to determine if the robot has been in the imaged location previously. Schubert \cite{Schubert2023VisualPR} presents a general software framework for evaluating a variety of image descriptors on a number of publicly available VPR datasets. Using this framework, we present a performance evaluation of the addition of novel synthesized views to these datasets.

Given a single camera view, it is possible to generate a novel viewpoint by geometric transformation of the original image \cite{hartley2004multiple}. However, this well-known methodology cannot reconstruct image information not present in the original image. Generative AI techniques fill in areas of the novel view not seen in the original image based on their training. 
GenWarp \cite{seo2024genwarp} is a diffusion-based system that aims to generate new camera views from a given image while maintaining scene semantics. It balances warping and generation to produce a result consistent with the original image and its context. 

We use GenWarp to produce  a range of novel views with a range of viewpoint changes from the reference and query images for several publicly available datasets and add the novel views to the datasets. We then use several state-of-the-art image descriptors to generate recall statistics.
These are compared to the  statistics for the unaltered datasets to understand if and how far novel view synthesis generates imagery that is consistent with real visual place imagery. 

The next section presents a review of the relevant generative visual AI literature and explains why GenWarp was selected. Section \ref{method} presents our experimental method and implementation. Section \ref{results} presents the result of the experimentation, and these are discussed in Section \ref{discussion}. Section \ref{conclusion} summarizes our conclusions and details next steps.

\section{Literature}

\noindent\underline{
Visual place recognition (VPR)} is the problem of recognizing when camera imagery indicates that a robot is at a location it has occupied before. VPR plays an important role in image retrieval, map closure, topological navigation, and multi-robot coordination \cite{zaffar2021vpr,Schubert2023VisualPR,lyonsrahouti2023}.
This is often characterized as an image retrieval process: given a reference database containing images of a set of places 
and a query image, find the place whose images best match the query image. Part of what makes this a challenging problem is that images of the same place may appear quite different due to seasonal changes, weather illumination, structural changes to the environment, transient pedestrian or vehicle traffic, and so forth \cite{Vysotska25}. 
A crucial part of the VPR problem is the processing of query and reference images to yield image descriptors, and there has been substantial research on this, including
CNN-based (AlexNet), VPR CNN-based (NetVLAD)\cite{arandjelovic2016netvlad} \cite{Berton2022RethinkingVG}, place-invariance \cite{Berton2023EigenPlacesTV},  \cite{mixvpr} and transformer-based  \cite{Zhu2023R2FU} methods. 
Given image descriptors, the reference images can be matched against the query images, yielding a similarity vector representing the match score for each image. 
VPR researchers present results in terms of metrics such as \textit{recall@K} and AUC \cite{Schubert2023VisualPR}. AUC (Area Under the Curve) measures the overall retrieval performance by computing the area under the precision–recall 
urve, summarizing how well a method distinguishes correct place matches from incorrect ones across all decision thresholds.

\noindent\underline{
Novel View Synthesis (NVS)} aims to produce photorealistic images of a scene from novel perspectives and has emerged as a critical challenge in 3D vision and robotics. Early methods used geometry-based methods, like multi-view stereo and image-based rendering, which needed precise depth or dense correspondences to distort images from different angles. Learning-based techniques like DeepStereo \cite{flynn2016deepstereo} brought convolutional architectures into the picture for view prediction. These studies showcase how data-driven models could get close to complicated geometric transformations without having to build a 3D model.

Neural Radiance Fields (NeRF)~\cite{mildenhall2021nerf} 
represent scenes as continuous volumetric functions allowing rendering with high fidelity from any camera position. Later  research developed NVS in a way that it became more useful for robotics by making it faster to render, more generalizable, and more scalable. More recently, light-weight warping-based methods like GenWarp \cite{seo2024genwarp} use estimated depth and camera transformations to quickly create new views without having to model the whole volume. These  are a good choice for real-time or large-scale experiments.

Novel view synthesis quality 
hasn't been fully explored as a controlled tool to improve 
robot place recognition and navigation.
In this paper, we systematically create novel   synthetic views with structured elevation and radial variation to evaluate recognition performance in a  visual place recognition pipelines. 

\begin{figure}
    \centering
    \includegraphics[width=3 in]{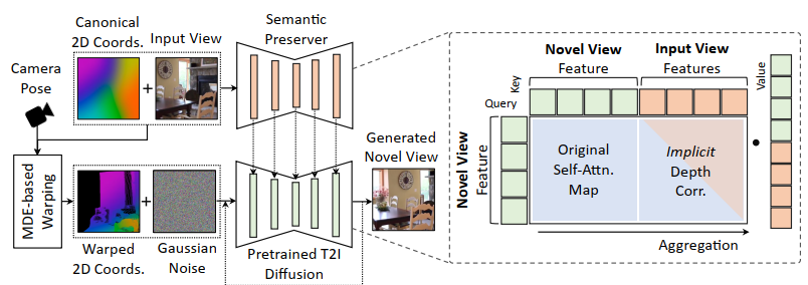}
    \caption{Model Architecture of the GenWarp framework (from \cite{seo2024genwarp}).}
    \label{fig:genwarp_MA}
\end{figure}

\section{Approach}\label{method}

The use of synthetic imagery to improve machine learning performance is not a new topic. Early work used procedurally generated images and simulation to expand datasets to improve performance \cite{Rozantsev2015Rendering}. State-of-the-art generative methods such as Stable Diffusion \cite{Rombach2022LDM} can generate large volumes of labeled and realistic imagery. Research has shown that augmenting datasets with this synthetic imagery can be effective in handling situations where there is insufficient or an imbalanced supply of real data or when the synthetic data is combined with real imagery \cite{Fanetal2024}.

\begin{figure}[ht]
    \centering
    \includegraphics[width=0.45\textwidth]{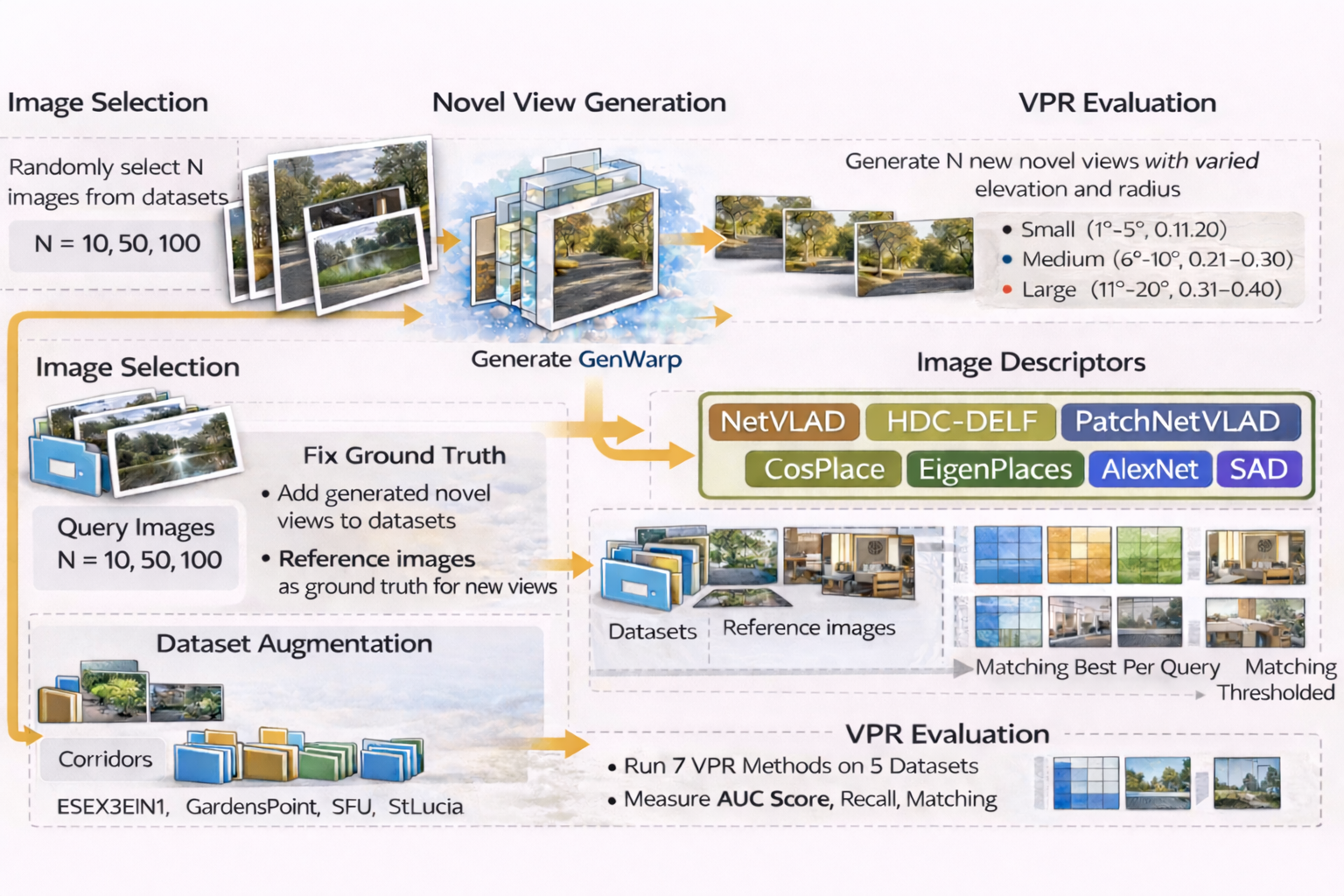}
    \caption{Methodology for Evaluating Injections of Synthetic Novels Views into VPR}
    \label{fig:method}
\end{figure}

Our principal interest is the potential of novel view synthesis for navigation: Generating a ground robot novel target view from an aerial view (or vice versa) or, in general, generating a novel expected view for one robot from an image taken by a second. As a first step to determine if this approach has value, we investigate the effect of adding novel synthetic views to a selection of VPR datasets and, using a selection of state-of-the-art image descriptors, evaluate the effect on the AUC metric (Fig. \ref{fig:method}). We argue that this is an appropriate approach because both VPR and our navigation objective have the same requirement: that query and reference views are of the same physical place from a different perspective or viewing condition. We acknowledge that VPR has a weaker requirement, since the query image just has to be visually consistent with the reference images, whereas for navigation, the novel synthetic view has to be sufficiently like the real, but previously unseen, view for navigation to be successful. Just performing well on our VPR evaluation won't guarantee the approach will work for navigation—but if it fails to work for our VPR evaluation, then it won't work for navigation. Thus, from our VPR experiment, we expect to learn something about the bounds within which the navigation approach might be expected to work.

Based on our intended application, we limit our interest to generative approaches that work from a single image \cite{seo2024genwarp,lin2023vision} rather than multiple images \cite{kerbl20233d,mildenhall2021nerf}. We selected GenWarp \cite{seo2024genwarp}, a diffusion-based system that aims to generate new camera views from a given image while maintaining scene semantics.

 \subsection{Genwarp} A number of prior warp-then-inpaint solutions heavily lean on monocular depth estimation (MDE). When the depth is noisy, these struggle to inpaint the unknown regions. GenWarp addresses these problems by integrating geometrical warping and generative synthesis into a single diffusion-based process so that the model can decide where to warp and where to generate.

GenWarp employs a two-stream architecture displayed in Fig. \ref{fig:genwarp_MA} which comprises:
\begin{itemize}
    \item \textbf{Semantic Preserver Network:} It extracts the semantic features from an input image by maintaining its contextual and structural information during novel view synthesis.
    \item \textbf{Diffusion U-Net:} Synthesizes the target novel view given depth, intrinsic parameters, and relative camera pose. The new {\bf viewpoint}  is given by spherical coordinates $(\phi,\psi,r)$ where $\phi$ and $\psi$ are azimuth and elevation and $r$ is distance relative to the original image.
\end{itemize}

The method enhances the self-attention mechanism in the diffusion U-Net with an adaptive cross-view attention, which enables the network to adaptively match source and target features. Such aggregated attention enables the model to balance geometric warping and generative synthesis. Along with that, it helps to perform well in image regions that are occluded or ambiguous in depth. A warped coordinate embedding encodes the geometric relationship between input and novel views by using estimated depth maps from ZoeDepth or DUSt3R. Unlike explicit pixel or feature warping, GenWarp achieves implicit geometric alignment through an attention mechanism, which helps avoid the propagation of depth.

The model is built on Stable Diffusion v1.5 and fine-tuned on large-scale multi-view datasets like RealEstate10K, ScanNet, and ACID. Training pairs are composed of source image $(I_i)$, target image $(I_j)$, monocular depth $(D_i)$ and relative camera pose $(P_{i \rightarrow j})$. For the datasets without ground-truth depth, pseudo-depths are generated by DUSt3R and aligned using PnP-RANSAC.



GenWarp qualitatively and quantitatively outperforms previous works such as GeoGPT, Photometric-NVS, and Stable Diffusion Inpainting. It shows better FID (visual quality) and PSNR (reconstruction accuracy) results on both in-domain (RealEstate10K) and out-of-domain (ScanNet) datasets. The synthesized views are held semantically consistent and have fewer depth-related artifacts, retaining realistic structure across a large change in viewing angle.

\subsection{Methodology}

Schubert \cite{Schubert2023VisualPR} provides a comprehensive tutorial on visual place recognition (VPR), outlining the core problem of recognizing previously visited locations under significant appearance, viewpoint, and environmental changes. The software for the VPR framework is publicly available on GitHub and forms the basis of our experimental methodology. The VPR Tutorial framework uses three public VPR datasets\footnote{https://www.tu-chemnitz.de/etit/proaut/datasets}: Gardenspoint, SFU and Santa Lucia. To this, we added two more datasets\footnote{https://github.com/MubarizZaffar/VPR-Bench/tree/main/datasets}: Corridor and ESSEX3IN1. We selected this collection of datasets because they cover a range of indoor and outdoor visual imagery. 

Each dataset was evaluated with seven state of the art image descriptors: NetVLAD\cite{arandjelovic2016netvlad}, HDC-DELF\cite{lee2020hdc}, PatchNetVLAD\cite{detone2021patchnetvlad}, CosPlace\cite{Berton2022RethinkingVG}, EigenPlaces\cite{Berton2023EigenPlacesTV}, AlexNet\cite{alexnet}, and SAD\cite{scharstein2002taxonomy}.
As a first step, we run these image descriptors on all five unaltered datasets and generate a base set of AUC statistics. These are shown in Table \ref{tab:auc_vpr} for each of the five datasets and for each of the seven image descriptors. This table will be the comparison as we inject synthetic views into the datasets.

\begin{table}[h]
\centering
\caption{Base Case: AUC Comparison of all Image Descriptors Across the Five Datasets}
\label{tab:auc_vpr}
\renewcommand{\arraystretch}{1.15}
\setlength{\tabcolsep}{5pt}
\begin{tabular}{lccccccc}
\hline
\textbf{Dataset} &
\textbf{Net} &
\textbf{HDC} &
\textbf{Patch} &
\textbf{Cos} &
\textbf{Eigen} &
\textbf{Alex} &
\textbf{SAD} \\
\hline
GardensPoint
& 0.199
& 0.740
& 0.802
& 0.593
& 0.675
& 0.195
& 0.030 \\

SFU
& 0.052
& 0.474
& 0.729
& 0.766
& 0.736
& 0.459
& 0.186 \\

StLucia
& 0.022
& 0.519
& 0.594
& 0.489
& 0.746
& 0.320
& 0.149 \\

Corridor
& 0.273
& 0.473
& 0.827
& 0.261
& 0.338
& 0.415
& 0.293 \\

ESSEX3IN1
& 0.532
& 0.112
& 0.961
& 0.861
& 0.875
& 0.063
& 0.033 \\
\hline
Avg
& 0.216
& 0.464
& 0.782
& 0.594
& 0.674
& 0.290
& 0.138 \\
\end{tabular}

\vspace{2pt}
{\footnotesize
Net = NetVLAD, 
HDC = HDC-DELF, 
Patch = PatchNetVLAD, 
Cos = CosPlace, 
Eigen = EigenPlaces, 
Alex = AlexNet.
}
\end{table}

\begin{figure}[ht]
    \centering
    \includegraphics[width=0.45\textwidth]{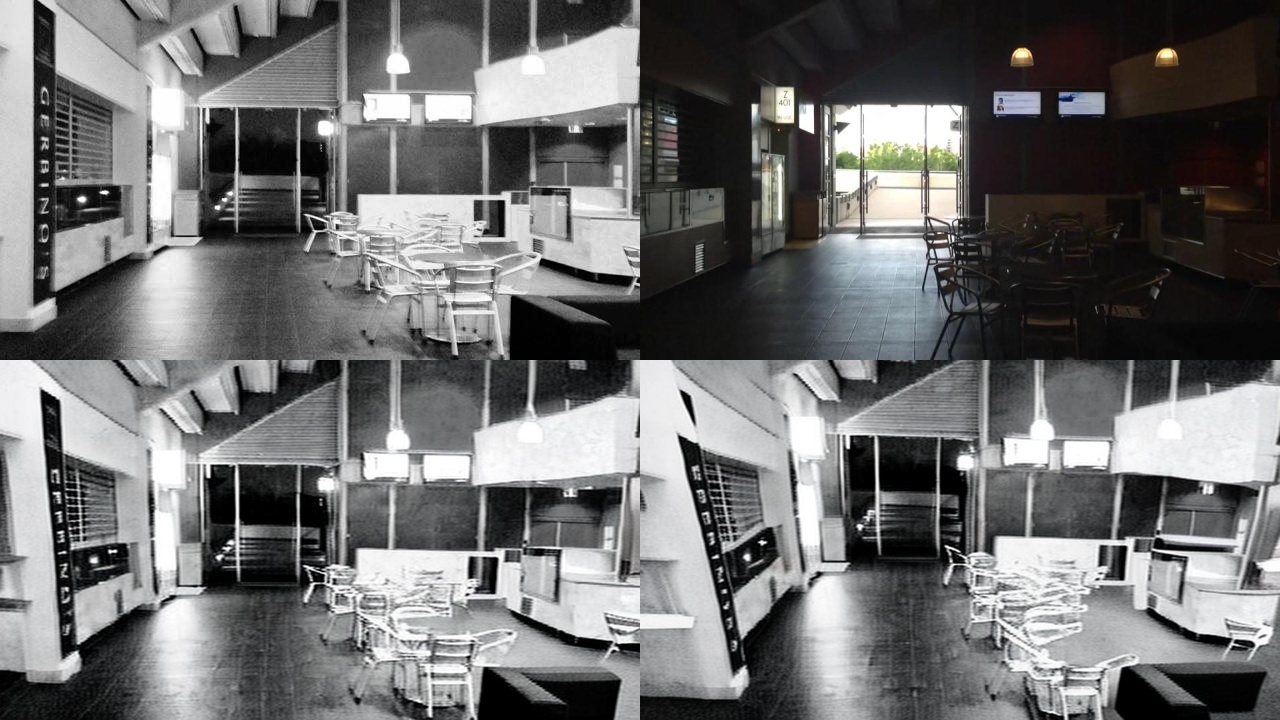}
    \caption{Example of Novel Synthetic Views generated for the GardensPoint VPR dataset. Clockwise from TL: Query Image,  Reference image, Novel Query with Medium elevation change, Novel Query with Large elevation change.}
    \label{fig:vpr_examples}
\end{figure}

\begin{table}[t]
\centering
\caption{Percentage of Dataset represented by injection of 10, 50 and 100 Synthetic Novel Views (Q: Query, R: Reference)}
\label{tab:dataset_inclusion}
\footnotesize
\setlength{\tabcolsep}{4pt}
\renewcommand{\arraystretch}{1.15}
\begin{tabular}{lcccc}
\hline
\textbf{Dataset} & \textbf{Q/R} & \textbf{+10 (\%)} & \textbf{+50 (\%)} & \textbf{+100 (\%)} \\
\hline
GardensPoint & 200/200 & 5 & 25 & 50 \\
SFU & 385/385 & 2.6 & 12.99 & 25.97 \\
StLucia & 200/200 & 5 & 25 & 50 \\
Corridor & 111/111 & 9 & 45.05 & 90.09 \\
ESSEX3IN1 & 210/210 & 4.76 & 23.81 & 47.62 \\
\hline
\end{tabular}
\end{table}

Our experimental methodology is as follows (see Fig. \ref{fig:method}):
\begin{enumerate}
    \item Randomly select $k$ images from the query set.
    \item Generate one novel synthetic view from each image at viewpoint $(\phi,\psi,r).$
    \item Set the ground truth for each novel view to be the same as for the original query image.
    \item Collect the AUC metric for all seven image descriptors for all five VPR datasets.
    \item Compare these results to the base results in Table \ref{tab:auc_vpr}.
\end{enumerate}

To evaluate how the quality of the results depends on the number of novel views added ($k$) and the magnitude of the change of viewpoint $(\phi,\psi,r)$, we look at three cases for each:
\begin{itemize}
    \item Amount of novel views injected, $k$: Small (10), Medium (50) and Large (100) injections of novel views.
    \item Viewpoint change of novel view $(\phi,\psi,r)$: 
    \begin{itemize}
    \item Small: $\phi,\psi\in\{0,\ldots,5^o\}, r\in\{0.01,\ldots,0.1\}$
    \item Medium: $\phi,\psi\in\{5,\ldots,10^o\}, r\in\{0.11,\ldots,0.2\}$
    \item Large: $\phi,\psi\in\{10,\ldots,20^o\}, r\in\{0.21,\ldots,0.3\}$
    \end{itemize}
\end{itemize}
Figure \ref{fig:vpr_examples} shows example imagery from the GardensPoint dataset and synthetic novel views for the Medium and Large viewpoint cases. 

Because the five datasets have different numbers of query/reference images, the impact of injecting 10, 50, and 100 novel views varies per dataset (Table \ref{tab:dataset_inclusion}). For example, the large injection (100 images) represents $50\%$ for GardensPoint but $90\%$ for the smaller Corridor dataset. Our results will need to be interpreted accordingly

\begin{table}[h]
\centering
\caption{AUC Comparison of all Image Descriptors Across the Five Datasets, Small injection (10) and  Small  viewpoint change.}
\label{tab:auc_vpr_low10_colored_merged}
\renewcommand{\arraystretch}{1.15}
\setlength{\tabcolsep}{5pt}
\footnotesize

\begin{tabular}{lccccccc}
\hline 
\multicolumn{8}{c}{\textbf{Generated Views Added to Query Set}} \\
\hline
\textbf{Dataset} &
\textbf{Net} &
\textbf{HDC} &
\textbf{Patch} &
\textbf{Cos} &
\textbf{Eigen} &
\textbf{Alex} &
\textbf{SAD} \\
\hline

GardensPoint
& \textcolor{green}{0.190}
& \textcolor{green}{0.744}
& \textcolor{green}{0.809}
& \textcolor{black}{0.582}
& \textcolor{black}{0.664}
& \textcolor{green}{0.200}
& \textcolor{green}{0.031} \\

SFU
& \textcolor{black}{0.022}
& \textcolor{green}{0.525}
& \textcolor{black}{0.729}
& \textcolor{red}{0.481}
& \textcolor{green}{0.734}
& \textcolor{red}{0.325}
& \textcolor{black}{0.146} \\

StLucia
& \textcolor{green}{0.047}
& \textcolor{red}{0.367}
& \textcolor{red}{0.457}
& \textcolor{green}{0.546}
& \textcolor{red}{0.518}
& \textcolor{green}{0.357}
& \textcolor{black}{0.136} \\

Corridor
& \textcolor{green}{0.267}
& \textcolor{black}{0.450}
& \textcolor{black}{\textbf{0.827}}
& \textcolor{black}{\textbf{0.261}}
& \textcolor{black}{0.319}
& \textcolor{black}{0.397}
& \textcolor{black}{0.273} \\

ESSEX3IN1
& \textcolor{black}{\textbf{0.532}}
& \textcolor{green}{0.115}
& \textcolor{black}{0.925}
& \textcolor{green}{0.870}
& \textcolor{black}{\textbf{0.875}}
& \textcolor{green}{0.057}
& \textcolor{green}{0.041} \\

\hline
Avg 
& 0.212
& 0.440
& 0.749
& 0.548
& 0.622
& 0.267
& 0.125\\

\multicolumn{8}{c}{\textbf{Generated Views Added to Reference Set}} \\
\hline

GardensPoint
& \textcolor{green}{0.189}
& \textcolor{black}{0.701}
& \textcolor{black}{0.779}
& \textcolor{black}{0.510}
& \textcolor{black}{0.604}
& \textcolor{green}{0.193}
& \textcolor{black}{\textbf{0.030}} \\

SFU
& \textcolor{black}{0.022}
& \textcolor{green}{0.517}
& \textcolor{green}{0.720}
& \textcolor{black}{0.681}
& \textcolor{green}{0.735}
& \textcolor{red}{0.317}
& \textcolor{black}{0.146} \\

StLucia
& \textcolor{green}{0.048}
& \textcolor{black}{0.460}
& \textcolor{green}{0.646}
& \textcolor{green}{0.532}
& \textcolor{red}{0.606}
& \textcolor{green}{0.353}
& \textcolor{black}{0.134} \\

Corridor
& \textcolor{green}{0.267}
& \textcolor{green}{0.466}
& \textcolor{green}{0.821}
& \textcolor{black}{0.237}
& \textcolor{black}{0.324}
& \textcolor{green}{0.408}
& \textcolor{green}{0.290} \\

ESSEX3IN1
& \textcolor{green}{0.588}
& \textcolor{green}{0.116}
& \textcolor{black}{0.930}
& \textcolor{green}{0.881}
& \textcolor{green}{0.884}
& \textcolor{green}{0.058}
& \textcolor{green}{0.045} \\

\hline
Avg
& 0.223
&  0.452
&  0.779
&  0.568
&  0.631
&  0.266
&  0.129\\

\end{tabular}

\vspace{2pt}
{\footnotesize
Metrics are marked green when they increased by any margin or decreased by not more than 0.009, marked black when they decreased by less than 0.09, marked red when they decreased by more than 0.09, and marked black in bold when they are unchanged compared with the base case.\\
Net = NetVLAD, 
HDC = HDC-DELF, 
Patch = PatchNetVLAD, 
Cos = CosPlace, 
Eigen = EigenPlaces, 
Alex = AlexNet.
}
\end{table}

Finally, we investigate whether it makes any difference to the performance whether the novel views are injected into the query set (keeping the reference set the same) or into the reference set (keeping the query set the same). The procedure for adding to the reference set is very similar to that described in more detail previously for the query set: an image is randomly selected from the reference set, a novel view of that image is generated, and ground truth is modified so that the new view has the same query as the original reference image.



\section{Results}\label{results}

Our results are presented as a series of tables that show the AUC metric for each dataset for each image descriptor, to be compared with the benchmark results for the unaltered datasets in Table \ref{tab:auc_vpr}. All the AUC values will be shown in each case; however, we establish the standard that 
metrics are marked green when they increased by any margin or decreased by not more than 0.009, marked black when they decreased by less than 0.09, marked red when they decreased by more than 0.09, and marked black in bold when they are unchanged compared with the base case. This allows us to get a quick sense of the results of the change.

\subsection{Query versus Reference injection results}

Table \ref{tab:auc_vpr_low10_colored_merged} shows the results when a small injection (10 views) with small viewpoint change is injected first into the query images and second into the reference images. We analyze the effect of this by comparing each metric to the corresponding metric for the unaltered dataset. We evaluate the effect numerically by calculating the average AUC per image descriptor across all the datasets (7 averages). These averages are shown as the last lines in Tables \ref{tab:auc_vpr} and \ref{tab:auc_vpr_low10_colored_merged}. 

The overall change with respect to the unaltered datasets is relatively small, from 1\% to 5\%, and mostly in the direction of improvement. The difference between adding the novel images to the reference set as opposed to the query set is also small, between 1\% and 3\%. 
While it's clear that the injection changes the AUC results, the change is small and is very similar for both cases. Based on this result, and to reduce the volume of results presented in this paper, we present the remaining results just for the case of injection of novel views into the query set and just for the medium and large change in viewpoint.

\subsection{Small, Medium and Large Injections}

Our main results are presented in Table \ref{tab:vpr_merged_colored} showing the small (10), medium (50), and large (100) injections of medium and large viewpoint change in all five datasets for all seven image descriptors. The green, red, and black color coding is present as before. It is immediately clear from the color coding that the AUC metric decreases as the number of files injected is increased. However, it can be seen that there does not appear to be much change for each injection as the viewpoint change is increased in magnitude.

The average metric for each image descriptor across all five datasets is calculated as before. The row labeled \enquote{Change} is the difference obtained by subtracting this average from the average for the image description in Table \ref{tab:auc_vpr} for the unaltered datasets.

\begin{table*}[t]
\centering
\caption{VPR Results Under Different Generated-Image Augmentation Settings (Compared to Base Case)}
\label{tab:vpr_merged_colored}
\footnotesize
\setlength{\tabcolsep}{3.2pt}
\renewcommand{\arraystretch}{1.15}

\resizebox{\textwidth}{!}{
\begin{tabular}{l|ccccccc|ccccccc}
\toprule
\multicolumn{15}{c}{\textbf{VPR RESULTS WITH 10 GENERATED-IMAGE AUGMENTATION}} \\
\toprule
\multirow{2}{*}{Dataset} &
\multicolumn{7}{c|}{Medium Viewpoint} &
\multicolumn{7}{c}{Large Viewpoint} \\
\cmidrule(lr){2-8} \cmidrule(lr){9-15}
& Net & HDC & Patch & Cos & Eigen & Alex & SAD
& Net & HDC & Patch & Cos & Eigen & Alex & SAD \\
\midrule

GardensPoint
& \textcolor{green}{0.198} & \textcolor{green}{0.739} & \textcolor{green}{0.808} & \textcolor{black}{0.583} & \textcolor{black}{0.663} & \textcolor{black}{\textbf{0.195}} & \textcolor{green}{0.031}
& \textcolor{green}{0.202} & \textcolor{black}{0.730} & \textcolor{green}{0.809} & \textcolor{black}{0.578} & \textcolor{black}{0.660} & \textcolor{green}{0.191} & \textcolor{black}{\textbf{0.030}} \\

SFU
& \textcolor{red}{0.022} & \textcolor{green}{0.519} & \textcolor{green}{0.727} & \textcolor{red}{0.482} & \textcolor{green}{0.735} & \textcolor{red}{0.322} & \textcolor{black}{0.147}
& \textcolor{red}{0.021} & \textcolor{green}{0.511} & \textcolor{green}{0.727} & \textcolor{red}{0.479} & \textcolor{green}{0.733} & \textcolor{red}{0.317} & \textcolor{black}{0.146} \\

StLucia
& \textcolor{green}{0.047} & \textcolor{red}{0.367} & \textcolor{red}{0.455} & \textcolor{green}{0.548} & \textcolor{red}{0.520} & \textcolor{green}{0.359} & \textcolor{black}{0.135}
& \textcolor{green}{0.047} & \textcolor{red}{0.365} & \textcolor{red}{0.454} & \textcolor{green}{0.546} & \textcolor{red}{0.518} & \textcolor{green}{0.358} & \textcolor{black}{0.135} \\

Corridor
& \textcolor{black}{0.268} & \textcolor{black}{0.443} & \textcolor{green}{0.828} & \textcolor{black}{0.249} & \textcolor{black}{0.311} & \textcolor{black}{0.388} & \textcolor{black}{0.273}
& \textcolor{black}{0.259} & \textcolor{black}{0.439} & \textcolor{black}{0.813} & \textcolor{black}{0.241} & \textcolor{red}{0.284} & \textcolor{black}{0.387} & \textcolor{black}{0.273} \\

ESSEX3IN1
& \textcolor{green}{0.568} & \textcolor{green}{0.117} & \textcolor{black}{0.922} & \textcolor{green}{0.871} & \textcolor{green}{0.877} & \textcolor{black}{0.058} & \textcolor{green}{0.042}
& \textcolor{green}{0.587} & \textcolor{green}{0.119} & \textcolor{black}{0.923} & \textcolor{green}{0.872} & \textcolor{green}{0.879} & \textcolor{black}{0.059} & \textcolor{green}{0.042} \\

\midrule
Avg
& 0.221
& 0.437	
& 0.748	
& 0.547	
& 0.621
& 0.264	
& 0.126
& 0.223
& 0.433
& 0.745
& 0.543
& 0.615
& 0.262
& 0.125\\
Change
& 0.005
&-0.027
&-0.035
& -0.047
& -0.053
& -0.026
& -0.013
& 0.008
& -0.031
&-0.037
& -0.051
& -0.059
& -0.028
& -0.013\\

\multicolumn{15}{c}{\textbf{VPR RESULTS WITH 50 GENERATED-IMAGE AUGMENTATION}} \\
\midrule

GardensPoint
& \textcolor{green}{0.190} & \textcolor{black}{0.723} & \textcolor{black}{0.782} & \textcolor{black}{0.544} & \textcolor{black}{0.625} & \textcolor{green}{0.193} & \textcolor{green}{0.032}
& \textcolor{black}{0.185} & \textcolor{black}{0.698} & \textcolor{black}{0.771} & \textcolor{black}{0.535} & \textcolor{black}{0.612} & \textcolor{black}{0.184} & \textcolor{black}{0.029} \\

SFU
& \textcolor{black}{0.021} & \textcolor{green}{0.499} & \textcolor{black}{0.701} & \textcolor{red}{0.444} & \textcolor{red}{0.682} & \textcolor{red}{0.305} & \textcolor{black}{0.147}
& \textcolor{black}{0.021} & \textcolor{green}{0.484} & \textcolor{black}{0.691} & \textcolor{red}{0.443} & \textcolor{red}{0.679} & \textcolor{red}{0.300} & \textcolor{black}{0.139} \\

StLucia
& \textcolor{green}{0.043} & \textcolor{red}{0.351} & \textcolor{red}{0.413} & \textcolor{green}{0.506} & \textcolor{red}{0.476} & \textcolor{green}{0.358} & \textcolor{black}{0.136}
& \textcolor{green}{0.041} & \textcolor{red}{0.336} & \textcolor{red}{0.407} & \textcolor{green}{0.501} & \textcolor{red}{0.469} & \textcolor{green}{0.343} & \textcolor{red}{0.125} \\

Corridor
& \textcolor{black}{0.250} & \textcolor{black}{0.442} & \textcolor{black}{0.794} & \textcolor{red}{0.157} & \textcolor{red}{0.186} & \textcolor{black}{0.390} & \textcolor{black}{0.274}
& \textcolor{black}{0.248} & \textcolor{black}{0.439} & \textcolor{black}{0.778} & \textcolor{red}{0.161} & \textcolor{red}{0.205} & \textcolor{black}{0.385} & \textcolor{black}{0.273} \\

ESSEX3IN1
& \textcolor{green}{0.551} & \textcolor{green}{0.111} & \textcolor{green}{0.970} & \textcolor{green}{0.911} & \textcolor{green}{0.905} & \textcolor{green}{0.062} & \textcolor{green}{0.045}
& \textcolor{green}{0.546} & \textcolor{black}{0.109} & \textcolor{black}{0.958} & \textcolor{green}{0.896} & \textcolor{green}{0.892} & \textcolor{black}{0.057} & \textcolor{green}{0.044} \\

\midrule
Avg
& 0.211	
& 0.425	
& 0.732	
& 0.512	
& 0.575	
& 0.262	
& 0.127
& 0.208&	0.413&	0.721&	0.507&	0.571&	0.254&	0.122\\

Change
& -0.005& 	-0.038&-0.051& 	-0.082& 	-0.099& 	-0.029& 	-0.012
& -0.007&	-0.051&	-0.062&	-0.087&	-0.103&	-0.037&	-0.016\\

\multicolumn{15}{c}{\textbf{VPR RESULTS WITH 100 GENERATED-IMAGE AUGMENTATION}} \\
\midrule

GardensPoint
& \textcolor{red}{0.168} & \textcolor{red}{0.625} & \textcolor{red}{0.640} & \textcolor{red}{0.478} & \textcolor{red}{0.547} & \textcolor{black}{0.158} & \textcolor{black}{0.023}
& \textcolor{red}{0.161} & \textcolor{red}{0.596} & \textcolor{red}{0.661} & \textcolor{red}{0.472} & \textcolor{red}{0.537} & \textcolor{black}{0.153} & \textcolor{black}{0.021} \\

SFU
& \textcolor{red}{0.018} & \textcolor{black}{0.418} & \textcolor{red}{0.566} & \textcolor{red}{0.392} & \textcolor{red}{0.597} & \textcolor{red}{0.253} & \textcolor{black}{0.120}
& \textcolor{red}{0.018} & \textcolor{black}{0.418} & \textcolor{red}{0.571} & \textcolor{red}{0.392} & \textcolor{red}{0.598} & \textcolor{red}{0.256} & \textcolor{black}{0.121} \\

StLucia
& \textcolor{green}{0.039} & \textcolor{red}{0.270} & \textcolor{red}{0.315} & \textcolor{black}{0.414} & \textcolor{red}{0.394} & \textcolor{black}{0.258} & \textcolor{black}{0.098}
& \textcolor{green}{0.040} & \textcolor{red}{0.274} & \textcolor{red}{0.317} & \textcolor{black}{0.417} & \textcolor{red}{0.395} & \textcolor{black}{0.252} & \textcolor{black}{0.102} \\

Corridor
& \textcolor{black}{0.220} & \textcolor{red}{0.302} & \textcolor{black}{0.764} & \textcolor{black}{0.201} & \textcolor{red}{0.238} & \textcolor{red}{0.279} & \textcolor{red}{0.141}
& \textcolor{black}{0.200} & \textcolor{red}{0.281} & \textcolor{black}{0.735} & \textcolor{black}{0.179} & \textcolor{red}{0.218} & \textcolor{red}{0.251} & \textcolor{red}{0.140} \\

ESSEX3IN1
& \textcolor{red}{0.433} & \textcolor{black}{0.086} & \textcolor{red}{0.695} & \textcolor{red}{0.659} & \textcolor{red}{0.664} & \textcolor{black}{0.048} & \textcolor{green}{0.035}
& \textcolor{red}{0.439} & \textcolor{black}{0.082} & \textcolor{red}{0.701} & \textcolor{red}{0.660} & \textcolor{red}{0.665} & \textcolor{black}{0.045} & \textcolor{black}{\textbf{0.033}} \\
\bottomrule
Avg 
& 0.176	
& 0.340
& 0.596
& 0.429
& 0.488	
& 0.199
& 0.083
& 0.172&	0.330&	0.597&	0.424&	0.483&	0.191&	0.083\\
Change
&-0.040&	-0.123&	-0.187&	-0.165&	-0.186&	-0.091&-0.055
&-0.044&	-0.133&	-0.186&-0.170&	-0.191&	-0.099&-0.055\\

\end{tabular}
}
{\footnotesize
Metrics are marked green when they increased by any margin or decreased by not more than 0.009, marked black when they decreased by less than 0.09, marked red when they decreased by more than 0.09, and marked black in bold when they are unchanged compared with the base case.\\
Net = NetVLAD, 
HDC = HDC-DELF, 
Patch = PatchNetVLAD, 
Cos = CosPlace, 
Eigen = EigenPlaces, 
Alex = AlexNet.
}
\end{table*}

\section{Discussion}\label{discussion}

\subsection{Effect of query versus reference injection}

The small change in metric that was recorded (Table  \ref{tab:auc_vpr_low10_colored_merged}) when a small amount of new views with small viewpoint change were injected into the five datasets is evidence in favor of the hypothesis that the novel views are compatible with the real scene geometry. 

It was observed that the change in metric was similar whether the injection was to the query dataset or the reference dataset, and that the change was in the direction of improvement over the unaltered datasets. We interpret this as further evidence that the novel views are compatible with the real scene geometry. They represent additional correct query matches on top of the unaltered dataset matches and hence improve the metric.

We  interpret these observations to mean novel synthetic views are indeed potentially useful for place recognition and navigation. However, the change in viewpoint  was very small in these experiments, certainly much less than would be needed in general in mapping views between a UAV and a ground robot.

\subsection{Effect of Viewpoint change}

The left and right sides of Table \ref{tab:vpr_merged_colored} show the effect of medium and large changes in viewpoint for the injected novel views. The average of averages for the medium change of viewpoint tables for each injection size are -0.028, -0.045, and -0.121 (top to bottom). For the large change of viewpoint tables, they are -0.030, -0.052, and -0.125. The difference in the corresponding averages (e.g., -0.028 and -0.030) is less than 1\% in each case. We argue. therefore, that the magnitude of the change in viewpoint has very little effect on performance up to the maximum angle size of $20^o$ and radius change of $0.3$.
This observation bodes well for the navigational use of a novel synthetic view generated from the view of one robot with the expected viewpoint of the second robot because we expect there will be a large change in viewpoint.

\subsection{Effect of Injection size}

Table 
\ref{tab:vpr_merged_colored} shows that the number of injected images degrades the AUC metric faster than the viewpoint change.
The two average of averages sequences discussed in the previous section  (-0.028, -0.045, -0.121 and -0.030, -0.052,  -0.125) shows an approx. 2\% decrease from 10 to 50 injected views and an 8\% decrease from 50 to 100 views. 

To look at these results more closely, we need to consider the relative impact of the injection size on the datasets, Table \ref{tab:dataset_inclusion}. Based on this, the impact on the Corridor dataset should be the greatest and on the SFU dataset should be the least, with the other three seeing somewhat similar effects. In fact, we find that the GardensPoint and ESSEX3IN1 datasets show the least effect, and StLucia shows the most effect. Our conclusion is that the nature of the imagery is more influential than the injection size. Arguably, GenWarp more effectively handles the simpler geometric scenes in GardensPoint and portions of ESSEX3IN1. The mixed natural/urban scenery in StLucia is more challenging. 

Looking at the image descriptor results more closely, SAD and NetVLAD are the least affected by the injected novel views. However, the metrics in each case were relatively low. 
The descriptor most impacted  was EigenPlaces, followed by CosPlace. 

The highest performing descriptor was PatchNetVLAD and the impact of the injection of novel views was in the middle of the range from SAD to Eigenplaces. Our conclusion based on this evaluation, limited as it is to just these five datasets, and the novel view generation viewpoint changes and injection amounts tested, is that PatchNetVLAD is best suited to image matching using novel synthetic views. 

\section{Conclusions}\label{conclusion}

We are interested in how novel synthetic view generation can be used in navigation, e.g., a novel overhead view generated from a scene taken by a ground robot could be used to guide an aerial robot to that location. To evaluate the feasibility of this approach, 
we use GenWarp to produce a gradated sequence of novel views from database and query images for five publicly available databases and use seven state-of-the-art image descriptors to generate recall statistics. These are compared to the statistics from the unaltered datasets to understand if and how far novel view synthesis generates imagery that is consistent with real visual place imagery. 

Our results are presented in several Tables but principally Table \ref{tab:vpr_merged_colored}. Our conclusions are summarized as follows:
\begin{itemize}
    \item Small injections (10) with a small change of viewpoint (angular change up to $5^o$ ) show a small improvement to the performance metric; we argue this is what we would expect if the novel views are consistent with additional real views. 
    \item Change of viewpoint magnitude (up to $20^o$) shows relatively little impact on performance for the larger injections of novel views. 
    \item The number of novel views added to the dataset can degrade performance by up to $8\%$. 
    \item The principal issue is not the percentage of the original dataset augmented with the new views, but rather the specific kind of imagery that is getting replaced. Imagery of corridors and buildings (GardensPoint, Corridor) was least affected, whereas more mixed scenery (StLucia) was affected more.
    \item The PatchNetVLAD image descriptor gave the result in terms of AUC value and tolerance of novel view injection.
\end{itemize}

This study involved only five datasets and a small scope for change of viewpoint. Future work will involve evaluating larger viewpoint changes and a wider variety of imagery.

\bibliographystyle{IEEEtran}
\bibliography{les}

\end{document}